# Answering Subcognitive Turing Test Questions: A Reply to French

Peter D. Turney
Institute for Information Technology
National Research Council of Canada
Ottawa, Ontario, Canada
email: peter.turney@nrc.ca
URL: http://purl.org/net/peter.turney

## Abstract

Robert French has argued that a disembodied computer is incapable of passing a Turing Test that includes *subcognitive* questions. Subcognitive questions are designed to probe the network of cultural and perceptual associations that humans naturally develop as we live, embodied and embedded in the world. In this paper, I show how it is possible for a disembodied computer to answer subcognitive questions appropriately, contrary to French's claim. My approach to answering subcognitive questions is to use statistical information extracted from a very large collection of text. In particular, I show how it is possible to answer a sample of subcognitive questions taken from French, by issuing queries to a search engine that indexes about 350 million Web pages. This simple algorithm may shed light on the nature of human (sub-) cognition, but the scope of this paper is limited to demonstrating that French is mistaken: a disembodied computer *can* answer subcognitive questions.

**Introduction**

Alan Turing (1950) proposed that we should consider a machine to be intelligent if it is capable of imitating human performance in a written, interactive test. This test has since become known as the Turing Test. Turing hoped that this operational test of machine intelligence would avoid the difficult and controversial problem of defining intelligence functionally or intensionally.

The Turing Test involves a human subject in one room, a machine subject in another room, and a human examiner in a third room. The examiner can communicate interactively with the two subjects by means of a teletype. The teletype is intended to screen the examiner from information that is not relevant to determining intelligence, such as the physical appearance of the subjects. The task of the examiner is to guess who is the human subject and who is not, by asking various questions to the two subjects and studying the replies. Both subjects are expected to make an effort to convince the examiner that they are the human subject. The machine is to be considered intelligent if the examiner cannot distinguish the two subjects more reliably than random guessing.

Robert French (1990, 2000) has argued that a disembodied computer cannot pass a Turing Test that includes *subcognitive* questions. He wrote (French, 2000):

> No computer that had not experienced the world as we humans had could pass a rigorously administered standard Turing Test. We show that the use of "subcognitive" questions allows the standard Turing Test to indirectly probe the human subcognitive associative concept network built up over a lifetime of experience with the world.

Subcognitive questions are designed to probe the network of associations that humans naturally develop, living embodied and immersed in the world. Subcognitive questions conform to the standard Turing Test, in that such questions and their answers can be produced with a teletype. Some examples will be presented later.

I am sympathetic to French's general position, that the Turing Test is too strong. It is a test of *human-like* intelligence, a test that a truly alien intellect would be bound to fail. Personally, as a researcher in Artificial Intelligence, I believe that it would be much more exciting to create a machine with non-human-like intelligence, than to create a machine with human-like intelligence. However, although I agree with French's position, I disagree with his argument in support of his position, his claim that a disembodied computer cannot imitate human answers to subcognitive questions. In the following, I will describe a simple unsupervised machine learning algorithm, called PMI-IR, that can generate human-like answers to subcognitive questions.

PMI-IR uses Pointwise Mutual Information (PMI) and Information Retrieval (IR) to measure the semantic similarity between pairs of words or phrases (Turney, 2001). The algorithm involves issuing queries to a search engine (the IR component) and applying statistical analysis to the results (the PMI component). The power of the algorithm comes from its ability to exploit a huge collection of text. In the following examples, I used the AltaVista® search engine, which indexes about 350 million Web pages in English.[1]

PMI-IR was designed to recognize synonyms. The task of synonym recognition is, given a problem word and a set of alternative words, choose the member from the set of alternative words that is most similar in meaning to the problem word. PMI-IR has been evaluated using 80 synonym recognition questions from the Test of English as a Foreign Language (TOEFL) and 50 synonym recognition questions from a collection of tests for students of English as a Second Language (ESL). On both tests, PMI-IR scores 74% (Turney, 2001). For comparison, the average score on the 80 TOEFL questions, for a large sample of applicants to US colleges from non-English speaking countries, was 64.5% (Landauer and Dumais, 1997). Landauer and Dumais (1997) note that, "… we have been told that the average score is adequate for admission to many universities." Latent Semantic Analysis (LSA), another statistical technique, scores 64.4% on the 80 TOEFL questions (Landauer and Dumais, 1997).[2]

PMI-IR is based on *co-occurrence* (Manning and Schütze, 1999). The core idea is that "a word is characterized by the company it keeps" (Firth, 1957). In essence, it is an algorithm for measuring the strength of associations among words, which makes it well suited for the task of answering subcognitive questions.

**Subcognitive Questions**

Here are some examples of subcognitive questions, taken from French (2000):

On a scale of 1 (awful) to 10 (excellent), please rate:

- How good is the name *Flugly* for a glamorous Hollywood actress?
- How good is the name *Flugly* for an accountant in a W.C. Fields movie?
- How good is the name *Flugly* for a child's teddy bear?

On a scale of 1 (terrible) to 10 (excellent), please rate:

- banana peels as musical instruments
- coconut shells as musical instruments
- radios as musical instruments

---

[1] The AltaVista search engine is a Web search service provided by the AltaVista Company of Palo Alto, California, http://www.altavista.com/. Including pages in languages other than English, AltaVista indexes more than 350 million Web pages, but these other languages are not relevant for answering questions in English. To estimate the number of English pages indexed by AltaVista, I used the Boolean query "the OR of OR an OR to" in the Advanced Search mode. The resulting number agrees with other published estimates. The primary reason for using AltaVista in the following examples is the Advanced Search mode, which supports more expressive queries than many of the competing Web search engines.

[2] This result for LSA is based on statistical analysis of about 30,000 encyclopedia articles. LSA has not yet been applied to text collections on the scale that can be handled by PMI-IR.

Please rate the following smells (1 = very bad, 10 = very nice):

- Newly cut grass
- Freshly baked bread
- A wet bath towel
- The ocean
- A hospital corridor

French (2000) wrote, regarding these questions:

> All of these questions attempt to elicit information from the vast, largely unconscious associative concept network that we have all built up over a lifetime of interacting with our environment. Furthermore, there is nothing "tricky" about these questions – for a human being, that is.
>
> It is worth considering this point in detail. Consider how good "Ethel Flugly" would be for the name for a glamorous Hollywood actress. It just doesn't work. (Anymore than "Archibald Leach" worked for a handsome male movie star … which is precisely why Hollywood movie moguls rechristened him "Cary Grant.") On the other hand, it works perfectly for an accountant in a W.C. Fields movie. Why? Because, in your mind's ear, you can hear a cantankerous W.C. Fields saying, "Flugly, get my gloves and let us pay a little visit to Miss Whipsnade." It also works for a child's teddy bear, because it partly activates words like "fluffy", "cuddly" (similar sounds), etc. Of course, "ugly" will become active but most likely in the sense of the Ugly Duckling, with all the connotations surrounding the loveable little duckling in the children's story, etc. In any event, even if we aren't sure exactly *why* it works, most people would agree that *Flugly* would be a downright awful name for a sexy actress, a good name for a character in a W.C. Fields movie, and a perfectly appropriate name for a child's teddy bear.

In the following, I will show that PMI-IR can generate *human-like* answers for the above examples of subcognitive questions. The answers to the three *Flugly* questions are consistent with French's intuitions, as described in the preceding quotation. An informal survey supports the claim that the answers to the remaining questions are also human-like. French himself seems to agree that the answers are human-like.[3]

**PMI-IR**

Consider the following synonym test question, one of the 80 TOEFL questions. Given the problem word *levied* and the four alternative words *imposed, believed, requested, correlated*, which of the alternatives is most similar in meaning to the problem word? Let *problem* represent the problem word and {$choice_1$, $choice_2$, …, $choice_n$} represent the alternatives. The PMI-IR algorithm assigns a score to each choice, score($choice_i$), and selects the choice that maximizes the score.

---

[3] Personal correspondence, June 7, 2001.

The PMI-IR algorithm is based on co-occurrence. There are many different measures of the degree to which two words co-occur (Manning and Schütze, 1999). PMI-IR uses Pointwise Mutual Information (PMI) (Church and Hanks, 1989; Church *et al.*, 1991), as follows:

$$\text{score}(choice_i) = \log_2(\text{p}(problem\ \&\ choice_i) / (\text{p}(problem)\text{p}(choice_i))) \qquad (1)$$

Here, p(*problem* & *choice_i*) is the probability that *problem* and *choice_i* co-occur. If *problem* and *choice_i* are statistically independent, then the probability that they co-occur is given by the product p(*problem*)p(*choice_i*). If they are not independent, and they have a tendency to co-occur, then p(*problem* & *choice_i*) will be greater than p(*problem*)p(*choice_i*). Therefore the ratio between p(*problem* & *choice_i*) and p(*problem*)p(*choice_i*) is a measure of the degree of statistical dependence between *problem* and *choice_i*. The log of this ratio is the amount of information that we acquire about the presence of *problem* when we observe *choice_i*. Since the equation is symmetrical, it is also the amount of information that we acquire about the presence of *choice_i* when we observe *problem*, which explains the term *mutual information*.[4]

Since we are looking for the maximum score, we can drop $\log_2$ (because it is monotonically increasing) and p(*problem*) (because it has the same value for all choices, for a given problem word). Thus (1) simplifies to:

$$\text{score}(choice_i) = \text{p}(problem\ \&\ choice_i) / \text{p}(choice_i) \qquad (2)$$

In other words, each choice is simply scored by the conditional probability of the problem word, given the choice word, p(*problem* | *choice_i*).

PMI-IR uses Information Retrieval (IR) to calculate the probabilities in (2). For the task of synonym recognition (Turney, 2001), I evaluated four different versions of PMI-IR, using four different kinds of queries. The following description of these four different methods for calculating (2) uses the AltaVista® Advanced Search query syntax.[5] Let hits(*query*) be the number of hits (the number of documents retrieved) given the query *query*. The four scores are presented in order of increasing sophistication. They can be seen as increasingly refined interpretations of what it means for two words to co-occur, or increasingly refined interpretations of equation (2).

---

[4] For an explanation of the term *pointwise* mutual information, see Manning and Schütze (1999).
[5] See http://doc.altavista.com/adv_search/syntax.html.

**Score 1:** In the simplest case, we say that two words co-occur when they appear in the same document:

$$\text{score}_1(choice_i) = \text{hits}(problem \text{ AND } choice_i) / \text{hits}(choice_i) \quad (3)$$

We ask the search engine how many documents contain both *problem* and *choice$_i$*, and then we ask how many documents contain *choice$_i$* alone. The ratio of these two numbers is the score for *choice$_i$*.

**Score 2:** Instead of asking how many documents contain both *problem* and *choice$_i$*, we can ask how many documents contain the two words close together:

$$\text{score}_2(choice_i) = \text{hits}(problem \text{ NEAR } choice_i) / \text{hits}(choice_i) \quad (4)$$

The AltaVista® NEAR operator constrains the search to documents that contain *problem* and *choice$_i$* within ten words of one another, in either order.

**Score 3:** The first two scores tend to score antonyms as highly as synonyms. For example, *big* and *small* may get the same score as *big* and *large*. The following score tends to reduce this effect, resulting in lower scores for antonyms:

$$\text{score}_3(choice_i) = \frac{\text{hits}((problem \text{ NEAR } choice_i) \text{ AND NOT } ((problem \text{ OR } choice_i) \text{ NEAR "not"}))}{\text{hits}(choice_i \text{ AND NOT } (choice_i \text{ NEAR "not"}))} \quad (5)$$

**Score 4:** The fourth score takes context into account. There is no context for the TOEFL questions, but the ESL questions involve context (Turney, 2001). For example, "Every year in the early spring farmers [tap] maple syrup from their trees (drain; boil; knock; rap)." The problem word *tap*, out of context, might seem to best match the choice words *knock* or *rap*, but the context *maple syrup* makes *drain* a better match for *tap*. In general, in addition to the problem word *problem* and the alternatives {*choice$_1$*, *choice$_2$*, ..., *choice$_n$*}, we may have context words {*context$_1$*, *context$_2$*, ..., *context$_m$*}. The following score includes a context word:

$$\text{score}_4(\textit{choice}_i) = \frac{\text{hits}((\textit{problem}\ \text{NEAR}\ \textit{choice}_i)\ \text{AND}\ \textit{context}\ \text{AND NOT}\ ((\textit{problem}\ \text{OR}\ \textit{choice}_i)\ \text{NEAR "not"}))}{\text{hits}(\textit{choice}_i\ \text{AND}\ \textit{context}\ \text{AND NOT}\ (\textit{choice}_i\ \text{NEAR "not"}))} \quad (6)$$

This equation easily generalizes to multiple context words, using AND, but each additional context word narrows the sample size, which might make the score more sensitive to noise (and could also reduce the sample size to zero). To address this issue, I chose only one context word from each ESL question. For a given ESL question, I automatically selected the context word by first eliminating the problem word (*tap*), the alternatives (*drain, boil, knock, rap*), and stop words (*in, the, from, their*). The remaining words (*every, year, early, spring, farmers, maple, syrup, trees*) were context words. I then used p(*problem* | *context*$_i$), as calculated by score$_3$(*context*$_i$), to evaluate each context word. In this example, *syrup* had the highest score (*maple* was second highest; that is, *maple* and *syrup* have the highest semantic similarity to *tap*, according to score$_3$), so *syrup* was selected as the context word *context* for calculating score$_4$(*choice*$_i$).

Table 1 shows how score$_3$ is calculated for the sample TOEFL question. In this case, *imposed* has the highest score, so it is (correctly) chosen as the most similar of the alternatives for the problem word *levied*.

Table 1. Details of the calculation of score$_3$ for a sample TOEFL question.

| Query | Hits |
|---|---:|
| imposed AND NOT (imposed NEAR "not") | 1,147,535 |
| believed AND NOT (believed NEAR "not") | 2,246,982 |
| requested AND NOT (requested NEAR "not") | 7,457,552 |
| correlated AND NOT (correlated NEAR "not") | 296,631 |
| | |
| (levied NEAR imposed) AND NOT ((levied OR imposed) NEAR "not") | 2,299 |
| (levied NEAR believed) AND NOT ((levied OR believed) NEAR "not") | 80 |
| (levied NEAR requested) AND NOT ((levied OR requested) NEAR "not") | 216 |
| (levied NEAR correlated) AND NOT ((levied OR correlated) NEAR "not") | 3 |

| Choice | | Score$_3$ |
|---|---:|---:|
| p(levied \| imposed) | 2,299 / 1,147,535 | 0.0020034 |
| p(levied \| believed) | 80 / 2,246,982 | 0.0000356 |
| p(levied \| requested) | 216 / 7,457,552 | 0.0000290 |
| p(levied \| correlated) | 3 / 296,631 | 0.0000101 |

As the scores increase in sophistication, their performance on the TOEFL and ESL questions improves. On the TOEFL test, score$_3$ gets 73.75% of the questions right (59/80) and on the ESL test, score$_4$ gets 74% of the questions right (37/50).[6] However, each increase in sophistication reduces the number of hits. For example hits(*problem* NEAR *choice$_i$*) will always be smaller than (or equal to) hits(*problem* AND *choice$_i$*). Thus each increase in sophistication is also an increase in the risk of random noise (the *sparse data problem*). Fortunately, this problem did not arise for the TOEFL and ESL tests, due to the large number of documents indexed by the search engine.

---
[6] Since the TOEFL questions have no context, score$_4$ cannot be used with them.

**Answering Subcognitive Questions with PMI-IR**

Before I begin answering the sample subcognitive questions, introduced above, I should discuss my methodology. Since the debate here is concerned with subcognition, rather than cognition, I assume that I am at liberty to use my cognitive powers, in order to transform the sample subcognitive questions into search engine queries and to analyse the results of the queries. That is, I will assume for the purposes of this argument that the hypothetical machine is capable of imitating human cognition, but not human subcognition.

There is a risk that I might abuse this assumption by transforming the subcognitive questions into queries, analysing the query results, and then "tweaking" the transformations until the results conform to my subcognitive expectations. Therefore I only permit myself one try for each question. The transformation from a subcognitive question to a rating from 1 to 10 is fixed in advance, before any search queries are issued, and cannot be changed, whatever the query results.

Let's begin with the three *Flugly* questions:

On a scale of 1 (awful) to 10 (excellent), please rate:

- How good is the name *Flugly* for a glamorous Hollywood actress?
- How good is the name *Flugly* for an accountant in a W.C. Fields movie?
- How good is the name *Flugly* for a child's teddy bear?

I think it is safe to say that it is cognitive (not subcognitive) knowledge that a question of the form, "How good is the name *X*?" should be answered by considering names that are similar in spelling to *X*. The AltaVista® Advanced Search query syntax includes an asterisk for matching alternative spellings.[7] For example, the query "colo*r" matches either "colour" or "color". An asterisk can match from 0 to 5 characters and must be preceded by at least three characters. Therefore the most general pattern available for finding words like *Flugly* is "Flu*". This pattern will only match a word that begins with a capital "F", which will tend to be either a name (proper noun) or the first word in a sentence.

It is also cognitive knowledge that we judge names by how familiar they are to us, or how similar they are to familiar names. This kind of question is different from a synonym recognition question, in that the important thing is the strength of the association; it does not matter whether the association is like a synonym or like an antonym. Therefore I will use a score that is like $score_4$, except that the NOT component, which was introduced for handling antonyms, will be dropped.

---
[7] See http://www.altavista.com/sites/help/search/search_cheat.

These considerations lead to the following transformations of the three questions:

$$p(\text{Flu*} \mid \text{actress}) = \frac{\text{hits}((\text{actress NEAR Flu*}) \text{ AND glamorous})}{\text{hits}(\text{actress AND glamorous})} \quad (7)$$

$$p(\text{Flu*} \mid \text{accountant}) = \frac{\text{hits}((\text{accountant NEAR Flu*}) \text{ AND movie})}{\text{hits}(\text{accountant AND movie})} \quad (8)$$

$$p(\text{Flu*} \mid \text{bear}) = \frac{\text{hits}((\text{bear NEAR Flu*}) \text{ AND teddy})}{\text{hits}(\text{bear AND teddy})} \quad (9)$$

The context words, *glamorous*, *movie*, and *teddy*, were chosen based on syntax. *Hollywood* and *W.C. Fields* have a capitalization pattern that indicates a proper noun. Proper nouns are generally less common than regular nouns and adjectives, therefore I avoided them, because they could result in the sparse data problem. I chose *teddy* rather than *child's*, because *child's* is a possessive form and *teddy* is an adjective, and also *teddy* immediately precedes *bear*, but *child's* is separated by one word, so it seemed that *teddy* was the more appropriate context for *bear*.

The equations, (7), (8), and (9), yield probability estimates ranging from zero to one, but the questions ask for a rating from one to ten. My methodology requires me to specify how the probability estimates will be mapped to a scale from one to ten, before I issue any search queries. I will use a linear mapping, where the smallest probability estimate maps to one, the largest probability estimate maps to ten, and intermediate probability estimates map linearly to values between one and ten.

Queries to the search engine yield the following results:

$$p(\text{Flu*} \mid \text{actress}) = 1 / 12{,}216 = 0.000082 \quad (10)$$

$$p(\text{Flu*} \mid \text{accountant}) = 4 / 21{,}682 = 0.00018 \quad (11)$$

$$p(\text{Flu*} \mid \text{bear}) = 421 / 508{,}833 = 0.00083 \tag{12}$$

When these probabilities are linearly mapped to the scale from one to ten, we have:

- *Flugly* for a glamorous Hollywood actress  = 1
- *Flugly* for an accountant in a W.C. Fields movie = 2
- *Flugly* for a child's teddy bear = 10

Recall that French (2000) wrote, "… most people would agree that *Flugly* would be a downright awful name for a sexy actress, a good name for a character in a W.C. Fields movie, and a perfectly appropriate name for a child's teddy bear." PMI-IR yields the same ranking as French (actress < accountant < bear). Perhaps French would give a higher score for Flugly as an accountant, but an informal survey suggests that the above ratings are quite human-like.

Let's examine the next three questions:

On a scale of 1 (terrible) to 10 (excellent), please rate:

- banana peels as musical instruments
- coconut shells as musical instruments
- radios as musical instruments

In this case, there are no context words. Again, it is cognitive knowledge that we judge music by familiarity, so we do not need to avoid antonyms. Therefore I use $\text{score}_2$ to transform these questions to queries:

$$p(\text{musical instruments} \mid \text{banana peels}) = \frac{\text{hits(musical instruments NEAR banana peels)}}{\text{hits(banana peels)}} \tag{13}$$

$$p(\text{musical instruments} \mid \text{coconut shells}) = \frac{\text{hits(musical instruments NEAR coconut shells)}}{\text{hits(coconut shells)}} \tag{14}$$

$$p(\text{musical instruments} \mid \text{radios}) = \frac{\text{hits(musical instruments NEAR radios)}}{\text{hits(radios)}} \tag{15}$$

Again, to map these probabilities to a scale from one to ten, I will use a linear mapping, where the lowest probability maps to one and the highest probability maps to ten.

These queries yield the following results:

$$p(\text{musical instruments} \mid \text{banana peels}) = 1 / 2{,}998 = 0.00033 \qquad (16)$$

$$p(\text{musical instruments} \mid \text{coconut shells}) = 5 / 1{,}880 = 0.0027 \qquad (17)$$

$$p(\text{musical instruments} \mid \text{radios}) = 1{,}253 / 1{,}006{,}207 = 0.0012 \qquad (18)$$

- banana peels as musical instruments = 1
- coconut shells as musical instruments = 10
- radios as musical instruments = 4

Again, an informal survey supports the claim that these results are human-like. There is some disagreement over the exact ratings, but there is consensus that the ranking is appropriate (banana peels < radios < coconut shells).

Finally, let's try the questions about smells:

Please rate the following smells (1 = very bad, 10 = very nice):

- Newly cut grass
- Freshly baked bread
- A wet bath towel
- The ocean
- A hospital corridor

I interpret these questions as asking whether a certain phrase is more semantically similar to *bad* or to *nice*, when the context word is *smell*. In this case, I want to avoid the antonyms *not bad* and *not nice*, so I will use score$_4$, including the NOT component. These queries are somewhat long, so I will only show the two equations for *newly cut grass*, as examples:

$$p(\text{nice} \mid \text{newly cut grass}) = \frac{\text{hits}((\text{newly cut grass NEAR nice}) \text{ AND smell AND NOT }((\text{newly cut grass OR nice}) \text{ NEAR "not"}))}{\text{hits}(\text{newly cut grass AND smell AND NOT }(\text{newly cut grass NEAR "not"}))} \qquad (19)$$

$$p(\text{bad} \mid \text{newly cut grass}) = \frac{\text{hits}((\text{newly cut grass NEAR bad}) \text{ AND smell AND NOT } ((\text{newly cut grass OR bad}) \text{ NEAR "not"}))}{\text{hits}(\text{newly cut grass AND smell AND NOT } (\text{newly cut grass NEAR "not"}))} \quad (20)$$

I use the following rules to map these probabilities into a scale from one to ten:

- If p(bad | *phrase*) = 0 and p(nice | *phrase*) = 0, then map *phrase* to 5.
- Otherwise, map *phrase* into the range from 1 to 10 using equation (21).

$$1 + 9 \times \left[ \frac{p(\text{nice} \mid phrase)}{p(\text{nice} \mid phrase) + p(\text{bad} \mid phrase)} \right] \quad (21)$$

The results of these queries are presented in Table 2.

Table 2. Query results for questions about smell.

| | | |
|---|---|---|
| p(nice | newly cut grass) | = 1 / 102 | = 0.0098 |
| p(bad | newly cut grass) | = 0 / 102 | = 0.0 |
| p(nice | freshly baked bread) | = 8 / 848 | = 0.0094 |
| p(bad | freshly baked bread) | = 0 / 848 | = 0.0 |
| p(nice | wet bath towel) | = 0 / 3 | = 0.0 |
| p(bad | wet bath towel) | = 0 / 3 | = 0.0 |
| p(nice | ocean) | = 270 / 45,360 | = 0.0060 |
| p(bad | ocean) | = 107 / 45,360 | = 0.0024 |
| p(nice | hospital corridor) | = 0 / 134 | = 0.0 |
| p(bad | hospital corridor) | = 0 / 134 | = 0.0 |

When we map these probabilities into the range from one to ten, we have the following answers:

- Newly cut grass         = 10
- Freshly baked bread     = 10
- A wet bath towel        =  5
- The ocean               =  7
- A hospital corridor     =  5

Once again, an informal survey suggests that these answers are human-like.

**Discussion**

The point of the previous section is to show that a disembodied computer can answer subcognitive questions in a human-like manner. When I have shown these results to various colleagues, they agree that the answers seem human-like, but they have raised other objections:

- The computer is not really disembodied, because it has access to the whole World Wide Web.
- An informal survey is not adequate. You need to ask a large number of people these questions, then see how far the computer deviates from the mean human response.
- The conversions of the subcognitive questions to search engine queries may be cognitive, but the choice of which conversions to make may have been influenced by subcognitive factors.
- Some of the hits that were generated by the queries may have been hits on French's paper or related papers, which could unfairly influence the results.

I will try to address each of these points.

A computer with access to the Web has indirect access to a vast amount of human experience, through the medium of the written word. However, I think it would be going too far to say that the computer is therefore *embodied*. This reminds me of a quotation, "People say life is the thing, but I prefer reading," attributed to Logan Pearsall Smith. As an avid reader, I am sympathetic to this point of view, but I would not agree that reading is an adequate substitute for having a body. Those who champion the *situated robotics* approach to artificial intelligence (e.g., Rodney Brooks, Randall Beer) would certainly not agree that a computer with Web access is thereby embodied. On the other hand, the results presented above do suggest that there is a lot of subcognitive knowledge latent in human writing. Perhaps there is a lesson here for situated robotics researchers, but that is tangential to the focus of this paper.

It is true that the above results would be more persuasive if I had done a survey of a large (human) population, in order to establish the mean and standard deviation of human replies to the sample subcognitive questions. However, I believe that the results, as they stand, are sufficient to cast doubt on French's (2000) claim that a disembodied computer would fail a Turing Test that includes subcognitive questions. In some sense, it could be argued that querying the Web is a kind of surrogate for surveying a large human population (which brings us back to the previous objection). Also, the work described here is a first effort at using a computer to answer subcognitive questions. If this first effort does not entirely succeed in emulating human replies, I think it at least suggests a research paradigm that could lead to more complete success. French's claim is about what is *possible*, in principle, and I think the results here are at least sufficient to show that it is *possible* for a disembodied machine to give human-like replies to subcognitive questions.

When I first tackled the subcognitive questions, I was aware of the possibility that my transformations of the questions into search engine queries might be influenced by subcognitive factors. This is why I adopted the methodology of restricting myself to only one try for each question. Otherwise, there would be too much risk that I might adjust the transformations until the results matched my own subcognitive expectations. However, it is possible that this methodology was only partially successful. It could be that I allowed subcognitive considerations to influence my transformations. Perhaps I subconsciously simulated the search engine queries in my mind and used these simulations to adapt the transformations. I can only say that I have no conscious awareness of this. To the best of my knowledge, the transformations were purely cognitive, not subcognitive. You may examine the transformations yourself (as presented in the previous section) to see whether you can detect a subcognitive component to them.

To address the objection that some of the search engine queries may have matched Web pages that discuss French's (1990, 2000) papers, I examined the hits in the cases where there was only a small number of hits, since these cases would be more susceptible to perturbation. None of these hits included French's papers or related work. For example, if you look at equations (7) and (10), you will see that the query "(actress NEAR Flu*) AND glamorous" had only one hit. This hit was a Web page discussing an advertising campaign by the National Fluid Milk Processors Promotion Board. The ads involved a "glamorous actress". The word "actress" appeared near the word "Fluid".

**Conclusion**

French (1990, 2000) has argued that the Turing Test is too strong, because a machine could be intelligent, yet still fail the test. I agree with this general point, but I disagree with the specific claim that an intelligent but disembodied machine cannot give human-like answers to subcognitive questions. I show that a simple approach using statistical analysis of a large collection of text can generate seemingly human-like answers to subcognitive questions.

The focus of this paper has been limited to addressing French's claim, but there may be more general lessons here about the nature of human cognition and the philosophical foundations of research in situated robotics and embodied artificial intelligence. However, I will leave this tangent for another paper.

**Acknowledgements**

Thanks to Arnold Smith, Joel Martin, Louise Linney, Norm Vinson, Michael Littman, Eibe Frank, David Nadeau, and Hassan Masum for helpful comments.